\documentclass[final]{cvpr}
\usepackage{times}
\usepackage{epsfig}
\usepackage{graphicx}
\usepackage{amsmath}
\usepackage{amssymb}
\usepackage{enumitem}
\usepackage{float}
\usepackage{dsfont}
\usepackage{isl-shortcuts}
\usepackage{tabularx}
\usepackage[accsupp]{axessibility}

\usepackage[pagebackref=true,breaklinks=true,colorlinks,bookmarks=false]{hyperref}

\def\cvprPaperID{5491} 
\def\confYear{CVPR 2022}

\begin{document}

\title{Dancing under the stars: video denoising in starlight}

\author{Kristina Monakhova\\
UC Berkeley\\
\and
Stephan R. Richter\\
Intel Labs \\
\and
Laura Waller\\
UC Berkeley \\
\and
Vladlen Koltun\\
Intel Labs \\
}

\makeatletter
\g@addto@macro\@maketitle{
	\begin{figure}[H]
		\setlength{\linewidth}{\textwidth}
		\setlength{\hsize}{\textwidth}
		\vspace{-7mm}
		\centering
		\includegraphics[width=\textwidth]{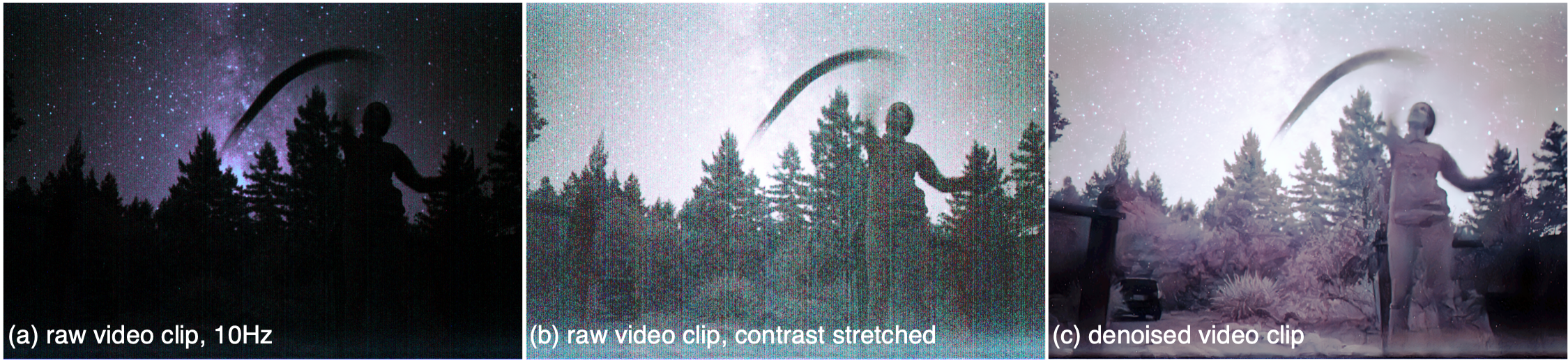}
        \caption{Video denoising in submillilux. (a) One frame from the raw noisy video clip (10 fps) taken between 0.6--0.7 millilux on a clear, moonless night with no external illumination. (b) Result after contrast stretching the video clip. (c) Denoised result using our denoiser.}
        \label{fig:short}
	\end{figure}
}
\makeatother

\maketitle
\thispagestyle{empty}

\begin{abstract}
Imaging in low light is extremely challenging due to low photon counts. Using sensitive CMOS cameras, it is currently possible to take videos at night under moonlight (0.05-0.3 lux illumination). In this paper, we demonstrate photorealistic video under starlight (no moon present, $<$0.001 lux) for the first time. To enable this, we develop a GAN-tuned physics-based noise model to more accurately represent camera noise at the lowest light levels. Using this noise model, we train a video denoiser using a combination of simulated noisy video clips and real noisy still images. We capture a 5-10 fps video dataset with significant motion at approximately 0.6-0.7 millilux with no active illumination.  Comparing against alternative methods, we achieve improved video quality at the lowest light levels, demonstrating  photorealistic video denoising in starlight for the first time. 
\vspace{-4mm}
\end{abstract}

\section{Introduction}
Some animals, such as hawkmoths and carpenter bees, can effectively navigate on the darkest moonless nights by the light of the stars ($<$ 0.001 lux)~\cite{somanathan2008visual, warrant2004vision, kelber2002scotopic}, while our best CMOS cameras generally require at least 3/4 moon illumination ($>$ 0.1 lux) to image moving objects at night~\cite{nightonearth}. Seeing in the darkest settings (moonless, clear nights) is extremely challenging due to the minuscule amounts of light present in the environment. In such dark settings, photographers can use long exposure times (e.g. 20 seconds or higher) to collect enough light from the scene. This approach works well for still images, but severely limits the temporal resolution and precludes imaging of moving objects. Alternatively, cameras can increase the gain, making each pixel effectively more sensitive to light. This allows shorter exposures, but greatly increases the noise present in each frame. In this setting, motion might be perceptible, but noise overwhelms the images.

Denoising algorithms can be used to improve the image quality in noisy images. Over the years, a number of denoising algorithms have been developed, from classic methods (e.g. BM3D~\cite{dabov2007image}, V-BM4D~\cite{maggioni2012video}) to deep learning-based approaches~\cite{zhang2017beyond}. Each of these methods attempt to extract the signal from the noise based on some assumptions about the statistical distributions of the image and noise. While successful for certain denoising tasks, most of these methods are built upon a simplistic noise model (Gaussian or Poisson-Gaussian noise), which does not apply in extremely low-light settings. When high sensor gain is used in low-light images, the noise is often non-Gaussian, non-linear, sensor-specific, and difficult to model or characterize. Without having a good understanding of the structure of the noise in the images, denoising algorithms may fail -- mistaking the structured noise for signal. 

\begin{figure*}[!t]
\centering
\includegraphics[width=\textwidth]{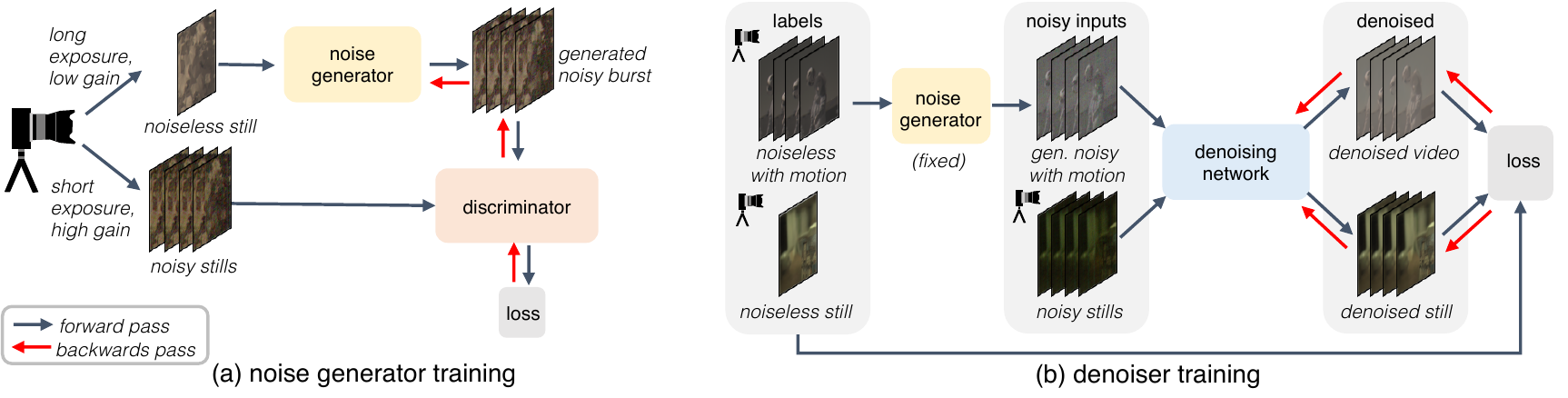}
   \caption{Method overview. (a) First we train our noise generator along with a discriminator, which aims to distinguish between real and synthetic noise. We use a limited dataset of long exposure/low gain and short exposure/high gain non-moving image pairs during this training process. After training, the noise generator can synthesize realistic noise. (b) Next, we train our denoiser using a combination of synthetic clean/noisy video clips produced using our noise generator as well as still clips from our camera. This allows us to train a video denoiser without needing experimental motion-aligned video pairs.}
\label{fig:overview}
\vspace{-4mm}
\end{figure*}

Recently, several deep learning-based approaches have provided remarkable denoising performance in low light down to 0.1--0.3 lux~\cite{chen2018learning, chen2019seeing}. Rather than assuming a certain noise model, these methods train a denoiser using clean/noisy image pairs captured by a camera. Such an approach automatically accounts for the low-light noise through deep learning, however this comes at the price of needing to capture thousands of training image pairs. Furthermore, the dataset is camera-dependent and must be retaken for each different sensor, since noise can be highly camera-specific. In addition, while it is possible to capture clean/noisy image pairs for non-moving objects by changing the exposure/gain settings, capturing clean/noisy image pairs for moving scenes adds additional complexities (e.g. needing a second camera, aligning motion), making this experimentally impractical~\cite{jiang2019learning}.

To achieve submillilux video denoising, we propose to use a combination of three things: 1) a very good camera optimized for low-light imaging and set to the highest gain setting (Sec.~\ref{sec:camera}), 2) learning our camera's noise model using a physics-inspired noise generator and easy-to-obtain still noisy images from the camera (Sec.~\ref{sec:generator}), and 3) using this noise model to generate synthetic clean/noisy video pairs to train a video denoiser (Sec.~\ref{sec:denoiser}). Since our physics-based noise generator is trained using a limited dataset of still clean/noisy bursts, we do not need to acquire experimental motion-aligned clean/noisy video clips, greatly simplifying the experimental setup and decreasing the amount of data we need to collect. After noise generator training, we hold the noise generator fixed and train our video denoising using a combination of still clean/noisy image bursts paired with synthetic video clips (Sec.~\ref{sec:denoiser}). Figure~\ref{fig:overview} summarizes this two-stage training approach for our noise generator and denoiser.

We demonstrate the effectiveness of our denoising network on 5-10 fps videos taken on a moonless clear night in 0.6 millilux, showing photorealistic video denoising in submillilux levels of illumination for the first time. We present several challenging scenes with extensive motion, in which subjects dance by only the light of the Milky Way as a meteor shower rains down from above. 



\section{Related work}

\noindent\textbf{Image and video denoising.}
A variety of techniques for image and video denoising have been proposed and studied throughout the years. Many of the classic denoising methods rely on specific image priors, such as sparsity~\cite{portilla2003image, elad2006image}, smoothness~\cite{rudin1992nonlinear}, or Gaussian mixture models~\cite{yu2011solving, dong2015image}. Others utilize a non-local strategy to collaboratively denoise similar patches across an image~\cite{buades2005non,lebrun2013nonlocal,dabov2007image,maggioni2012video}. More recently, deep learning-based approaches have been applied to image denoising, in which an image prior is learned from the data rather than explicitly assumed~\cite{burger2012image,ehret2019model,claus2019videnn, tassano2020fastdvdnet, vaksman2021patch}. These methods have shown significant improvements over classic methods in terms of image quality, however they often make simplistic assumptions on the noise statistics, such as i.i.d Gaussian. When trained with these simplistic assumptions, classic techniques such as BM3D often outperform deep learning-based methods on real photographs with real noise~\cite{plotz2017benchmarking}. In this regard, several datasets of noisy and clean image pairs from real cameras have been created to benchmark and improve the performance of deep learning-based denoisers on real cameras~\cite{anaya2018renoir,plotz2017benchmarking, abdelhamed2018high}. In addition, some work has focused on ``unprocessing" online image datasets to better match RAW image distributions in order to generate more synthetic data for training RAW image denoisers~\cite{brooks2019unprocessing}.



Alternatively, another line of deep learning-based denoising focuses on unsupervised learning, in which no ground truth images are used to train the denoiser. Such methods either assume that the structure of a deep network can act as a prior for the image denoising ~\cite{ulyanov2018deep}, or assume statistical independence of the noise to train a denoiser using samples drawn from one~\cite{batson2019noise2self, krull2019noise2void, krull2020probabilistic, prakash2020fully} or multiple~\cite{lehtinen2018noise2noise} noisy image frames. While this line of work is promising since it does not require ground truth data and therefore can be adapted for different camera sensors, these methods are not easily adaptable to the more structured or signal-dependent noise that is present in low-light settings under high gain, such as banding noise.

\vspace{1mm}\noindent\textbf{Low-light photography.}
A number of methods focus particularly on the challenging case of denoising for low-light and night photography. A popular method for low-light photography is burst denoising, in which multiple images are merged and denoised, as in HDR+ and Google Night Sight~\cite{ hasinoff2016burst,liba2019handheld}. These methods require robust alignment techniques to account for any motion in the scene, which is difficult in the presence of extreme noise. A number of approaches have emerged that attempt to do this burst-alignment step automatically through deep learning~\cite{mildenhall2018burst, godard2018deep}. Often, the final goal of burst denoising is to obtain a single clean image from the noisy burst. In our work, we aim to obtain a full denoised video rather than a single clean image. 

Recently, a number of deep learning-based methods attempt to address low-light photography by learning to denoise images in the presence of extreme noise. These methods learn a denoiser and image enhancer network for underexposed low-light images by first collecting a training dataset of clean/noisy image pairs~\cite{chen2018learning, jiang2019learning,chen2019seeing} and have demonstrated remarkable results for low-light conditions down to 0.1~\si{\lux}. 
Our work pushes this limit down by two orders-of-magnitude, demonstrating video denoising below 1~\si{\milli\lux}.
Furthermore, these methods rely on a camera-specific dataset of ground truth/noisy image pairs for training, which is particularly challenging for video denoising. Our approach only requires a limited dataset of still image pairs for video denoising, eliminating the need for a large experimental dataset of noisy/clean aligned videos.



\vspace{1mm}\noindent\textbf{Noise models.}
A Gaussian noise model is commonly used for typical imaging systems, however this is not a very realistic representation of real-world sensor noise~\cite{plotz2017benchmarking}. Signal-dependent models, such as Poisson-Gaussian~\cite{foi2008practical, foi2009clipped} or a heteroscedastic Gaussian model~\cite{hasinoff2014photon} are more realistic as they account for the effect of shot noise in cameras. However, there are many more effects, such as clipping~\cite{foi2008practical}, fixed pattern noise, and banding noise that these models don't account for~\cite{konnik2014high, boukhayma2018low}. Additional work has focused on characterizing sensor noise in low-light environments by fitting to certain distributions for different noise components~\cite{wang2019enhancing, wei2020physics}. In general, noise modeling for extremely low-light imaging is complicated and it is difficult to accurately characterize and synthesize realistic camera noise, since the noise can be highly structured and sensor-dependent~\cite{emva1288, konnik2014high}. 

Recently, rather than characterizing the sensor noise, several methods have attempted to learn to synthesize realistic noise using generative adversarial networks (GANs)~\cite{chen2018image, tran2020gan} and normalizing flow models~\cite{abdelhamed2019noise}. However, physics-based statistical methods tend to outperform DNN-based methods~\cite{zhang2021rethinking}. We combine physics-based statistical methods with GAN-based training techniques to learn to approximate the sensor noise in a data-driven manner, without the need for hand-calibrating a noise model.

\section{Physics-inspired noise generator}
\label{sec:generator}
Cameras aim to exactly measure and record the light intensity of a scene, converting photons to voltage readings, which are then converted to bits by an analog to digital converter (ADC). Throughout this process, noise is inadvertently added to the measurement both as a function of photon statistics and the circuitry of the sensor. In well-lit environments, sensor noise is well understood and can be modeled as a combination of two primary components: shot noise, which originates from photon arrival statistics, as well as read noise, which is caused by imperfections in the sensor readout circuitry~\cite{hasinoff2014photon}. In low-light settings, this noise approximation breaks down and does not adequately describe the complex noise statistics of the scene. Previous work has shown that noise in low-light settings can be expressed as a combination of photon shot noise, read noise, row noise, and quantization noise, which can be estimated through a rigorous calibration process~\cite{wei2020physics}. Inspired by this work, we propose a physics-inspired noise generator which consists of several learned statistical noise parameters. Rather than calibrating the noise parameters by hand, we automatically learn the optimal parameters using a GAN that is fed several pairs of calibration clean (long exposure, low gain)/noisy (short exposure, high gain) image pairs. Using this framework, our noise generator is trained to synthesize realistic noise in extremely low light and high gain settings, see Fig.~\ref{fig:noise_model}.

\subsection{Physics-inspired parameters}

\begin{figure}[thb]
\centering
\includegraphics[width=\linewidth]{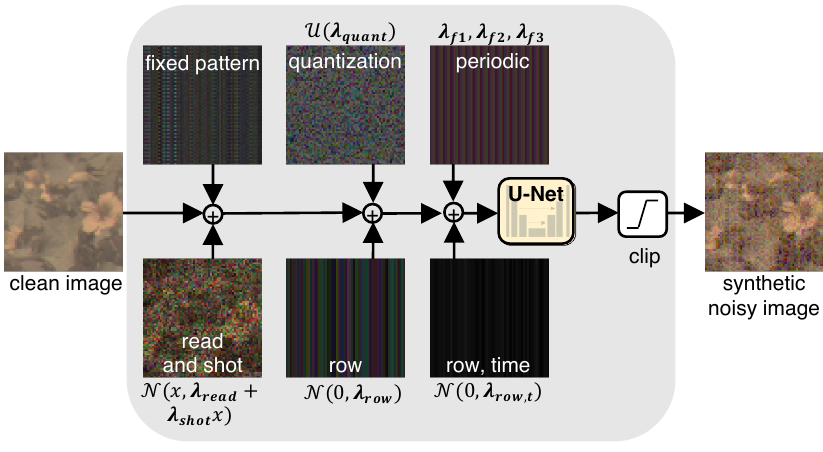}
   \caption{Physics-inspired noise generator. Our noise generator takes in a clean image and produces a synthetic noisy image. During training, our physics-inspired statistical noise parameters are optimized along with a U-Net to produce a synthetic noisy image that is indistinguishable from a real noisy image.}
\label{fig:noise_model}
\vspace{-3mm}
\end{figure}

Our noise generator contains several physics-inspired parameters, mainly consisting of variance terms for random distributions, as well as a convolutional neural network (CNN) to capture any additional effects, known or unknown, that are difficult to specifically model. Both components are jointly optimized during training. First, we parameterize the contributions of read and shot noise. Shot noise is a function of the light intensity hitting the sensor and is often modeled as a Poisson random variable, whereas read noise can be approximated as a zero-mean Gaussian random variable~\cite{foi2008practical, foi2009clipped}. Together, these are commonly approximated using a single heteroscedastic Gaussian random variable, where the mean is equal to the true signal, $x$, and the variance is parameterized by the read, $\lambda_{read}$, and shot noise, $\lambda_{shot}$. We note that a Poisson noise model is more accurate for low photon counts, but a Gaussian model is differentiable with respect to its mean and variance, allowing these parameters to be learned:
\begin{equation}
   N_{s} + N_{r} \sim \mathcal{N} (\mu = x, \sigma^2 = \lambda_{read} + \lambda_{shot}x)
\end{equation}

Low-light imaging often suffers from banding noise, which is a camera-dependent noise that results from the camera circuitry and is particularly prominent at high ISO settings. Banding noise often appears as horizontal or vertical lines in the measurement~\cite{konnik2014high, boukhayma2018low}. We model this as a fixed offset added to each column/row, where the fixed offset is drawn from a zero-mean Gaussian random variable with variance $\lambda_{row}$, see Fig.~\ref{fig:noise_model}. 
Banding noise is generally independent for each frame, however we have noticed that in extreme low light and high gain settings some banding patterns are consistent across a number of frames. To model this, we also include a time-consistent banding pattern noise which is static across each set of frames. As with the original banding noise, this time-consistent noise is modeled as a zero-mean Gaussian random variable with variance $\lambda_{row, t}$. 

In addition to this, at extreme gain settings, we notice that the measurements suffer from periodic noise, potentially due to ADC imperfections/effects at these high gain settings. This periodic noise appears as spikes in the frequency domain of the raw noisy measurements, corresponding to adding a 1 or 2 pixel period sinusoidal pattern to the image with a random amplitude (Fig.~\ref{fig:noise_model}). We parameterize this random amplitude by learned parameters: $\lambda_{f1}$, $\lambda_{f2}$, $\lambda_{f3}$. See Suppl.\ for further implementation details and discussion. 


Next, we add a uniform noise component, to approximate quantization noise in the sensor:
\begin{equation}
N_{q} \sim \mathcal{U}(\lambda_{quant}).
\end{equation}
Here $\lambda_{quant}$ is our parameter for the quantization noise interval. Generally, this noise component is well-defined based on the number of bits used by the camera sensor. However, we find that allowing this noise parameter to vary can improve our noise generator. Finally, we include a fixed pattern noise component, $N_{f}$, that stays constant throughout all images. We measure this experimentally using an average of several image sequences. We find that letting this fixed pattern noise be learned can improve the Kullback–Leibler (KL) divergence between the real noise and generated noise, but this parameter is prone to overfitting and we achieve the best denoising performance when leaving $N_{f}$ fixed and experimentally measured. 

Thus, our physics-inspired noise model consists of the following components:
\begin{equation}
    N = N_{s} + N_{r} + N_{row} + N_{row,t} + N_{q} + N_{f} + N_{p},
\end{equation}
where $N_{shot}, N_{read}, N_{row}, N_{row,t}, N_{q}, N_{f},$ and $N_{p}$ approximate the contributions of shot noise, read noise, row noise, temporal row noise, quantization noise, fixed pattern, and periodic noise.

After initial noise is added to a clean image using the physics-inspired parameters, the intermediate noisy image is passed through a CNN, which aims to improve the initial noise estimate and capture any effects that were not captured by the physics-inspired noise model. We utilize a residual 2D U-Net~\cite{ronneberger2015u} for this. (See Suppl.\ for architecture details.)
The final output of our noise generator is clipped to $[0,1]$.
Together, we have a total of 8 physics-inspired parameters ($\lambda_{read}$, $\lambda_{shot}$, $\lambda_{quant}, \lambda_{row}, \lambda_{row_t}, \lambda_{f1}, \lambda_{f2}, \lambda_{f3}$), as well as the parameters of the U-Net. All parameters are optimized during training to produce a realistic synthetic noisy image from a noiseless image. Figure~\ref{fig:noise_model} shows our physics-guided noise generator with a sample for each noise component.

\subsection{GAN training}
We want our noise generator to produce different noise samples at each forward pass. This is incompatible with direct supervision where each clean image would be paired to a ground truth noisy image.
Thus, to train our noise generator, we resort to an adversarial setup~\cite{goodfellow2014}, consisting in this case of our noise generator and a discriminator, which evaluates the realism of synthesized noisy images.

Our discriminator operates on noise patches of size 64x64. For our training objective, we utilize a standard Wasserstain GAN with a gradient penalty framework~\cite{gulrajani2017improved}, which is optimized with  the following objective function:
\begin{equation}
    L = \underset{\tilde{x} \sim \mathbb{P}_g}{\mathbb{E}} [D(\tilde{x})] - \underset{x \sim \mathbb{P}_r}{\mathbb{E}} [D(x)] + \lambda \underset{\hat{x} \sim \mathbb{P}_{\hat{x}}}{\mathbb{E}} \| (\nabla_{\hat{x}} D(\hat{x})\|_2 - 1)^2], 
\end{equation}
where $\mathbb{P}_r$ is the real noisy data distribution, $\mathbb{P}_g$ is the model distribution defined by the generator, $\tilde{x} = G(z)$, $z$ is a noiseless image patch, and $D$ is our discriminator. See Suppl.\ for full training details.



\section{Camera selection and data collection}
\label{sec:camera}
To meet our objective of producing photorealistic videos at submillilux illumination levels, we need to carefully choose an appropriate camera sensor and lens. Generally, larger pixel sizes are preferable for low-light imaging, so that each pixel can collect more photons. In addition, near infrared (NIR) sensitivity is useful for nighttime imaging because there are more detectable photons in NIR than at RGB wavelengths at night~\cite{leinert19981997, vollmerhausen2003night}.

We choose to use a Canon LI3030SAI Sensor, which is a 2160x1280 sensor with 19$\mu$m pixels, 16 channel analog output, and increased quantum efficiency in NIR. This camera has a Bayer pattern consisting of red, green, blue (RGB), and NIR channels (800-950nm). Each RGB channel has an additional transmittance peak overlapping with the NIR channel to increase light throughput at night. During daylight, the NIR channel can be subtracted from each RGB channel to produce a color image, however at night when NIR is dominant, subtracting out the NIR channel will remove a large portion of the signal resulting in muffled colors. We pair this sensor with a ZEISS Otus 28mm f/1.4 ZF.2 lens, which we choose due to its large aperture and wide field-of-view. 

We capture 3 sets of datasets from our camera: bursts of paired clean (low gain, long exposure)/noisy (high gain, short exposure) static scenes, clean videos of moving objects, and noisy videos of moving objects in submillilux conditions. All images/videos are captured in RAW format. The paired dataset of static scenes is used to train our noise generator. Both the paired dataset and the clean videos of moving objects are used to train the denoiser. The final dataset is reserved for testing the performance of our denoiser in the most challenging setting. Our submillilux dataset can be used as a challenge for future denoising algorithms~\cite{starlightdataset}.

\vspace{1mm}\noindent\textbf{Paired clean/noisy bursts of static scenes.}
We collect 10 clips of grayscale and color targets, each with one clean image along with a burst of 100-900 noisy images, resulting in a dataset with 2558 noisy images.  We use this dataset exclusively for the noise generator training. In addition to this, we acquire a more complex dataset of 67 clean/noisy image pairs consisting of 16 noisy bursts for each clean image. This second dataset contains indoor and outdoor scenes with various lighting conditions. We use this dataset both for our noise generator and denoiser training. 

\vspace{1mm}\noindent\textbf{Unpaired clean RGB+NIR videos.}
With our trained noise generator, we can generate unlimited amounts of clean/noisy pairs from clean videos. Given an absence of open-sourced RGB+NIR RAW datasets, we collect our own dataset of noiseless video clips (unpaired). We collect 10 video sequences which we break into 166 video clips for training and 10 for testing. The videos are taken at different frame rates of both indoor and outdoor scenes. We capture these images at a low-gain setting during the daytime.

To augment our dataset, we utilize 329 video clips from the MOT video challenge~\cite{leal2015motchallenge}, which we then unprocess~\cite{brooks2019unprocessing} to resemble RAW video clips. While these video clips have significant motion, they have a different distribution of colors than the raw data from our camera. Due to this, we utilize the MOT videos during our initial pre-training step, then refine our denoiser using only the video clips from our camera. 

\vspace{1mm}\noindent\textbf{Submillilux RGB+NIR videos.}
To test our method in the lowest light settings, we collected videos in a remote location on a clear, moonless night (outside of most of the night-glow from cities). Throughout our experiments, no outside light sources were used to illuminate the scene. The illuminace, measured by a PR-810 Prichard Photometer, ranged within 0.6-0.7~\si{\milli\lux}, which is within the range expected for a clear moonless night. Videos were taken with exposures ranging within 0.1-0.2\si{\milli\second} per image, corresponding to 10-5fps. All videos were taken with the largest lens aperture to maximize the amount of light hitting the sensor, and at the highest gain settings for the camera.

\section{Video Denoising}
\label{sec:denoiser}
Now that we can generate clean/video pairs, our next step is to train a denoiser that will generalize well to real noisy video clips from our camera. Inspired by burst denoising, in which a burst of multiple noisy frames are used together to denoise a central frame, we choose a network architecture that can operate on multiple frames at a time. This is beneficial because denoising a burst of images can improve PSNR over single-image denoising, especially in photon-starved regimes.  In addition, video denoising can help us maintain temporal consistency across frames and reduce flickering throughout the denoised video. 

\begin{figure}[thb]
\centering
\includegraphics[width=\linewidth]{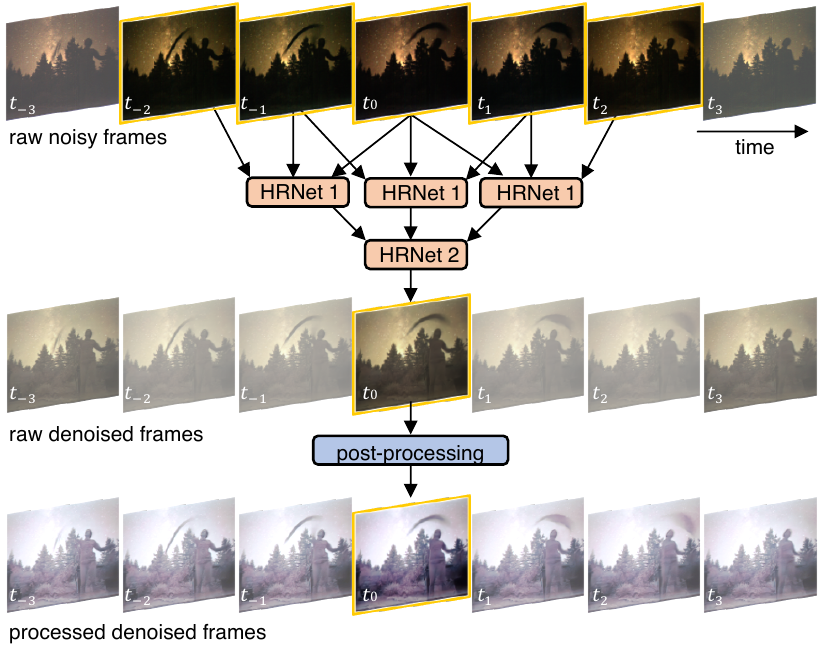}
   \caption{Denoising network. Our denoising network has a similar overall structure to FastDVDnet~\cite{tassano2020fastdvdnet}, sequentially taking in 5 noisy RAW images to produce 1 denoised RAW image. After denoising, off the shelf post-processing (e.g.\ white-balancing, histogram equalization) is applied to produce the final denoised video. }
\label{fig:denoiser}
\vspace{-4mm}
\end{figure}

\begin{figure*}[!thb]
\centering
\includegraphics[width=\linewidth]{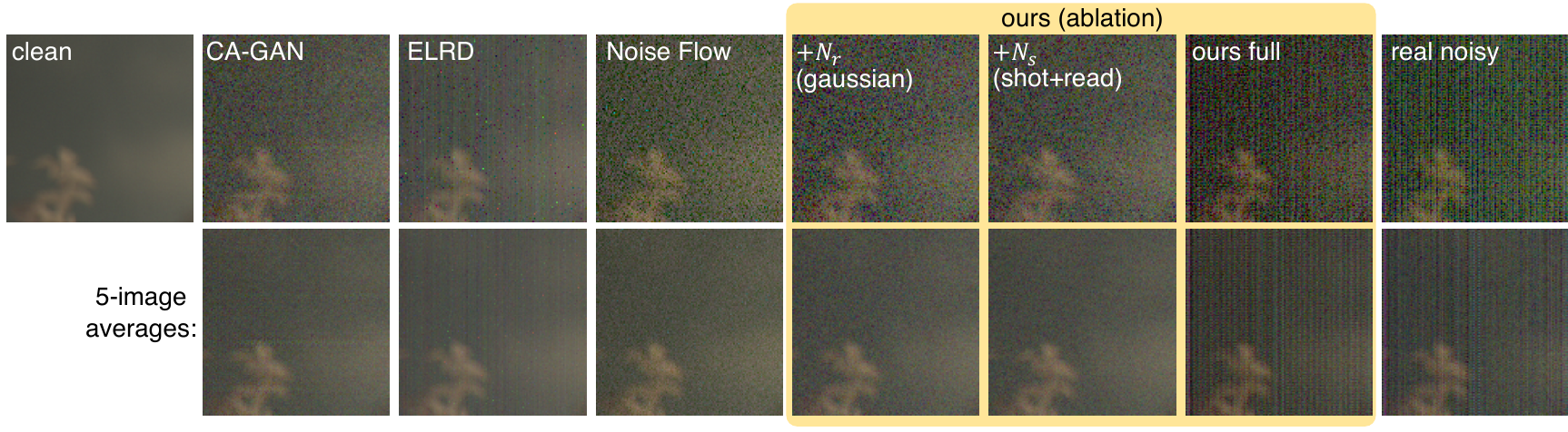}
   \caption{Noise model comparison. We show example image patches with our noise model vs.\ alternative noise models, as well as the mean of this image patch over 5 samples. Our synthetic noise appears more similar to the real noise than alternative methods and closely matches the average noise as well.}
\label{fig:noise_comparison}
\vspace{-4mm}
\end{figure*}

\subsection{Denoising Network}
For our denoising network, we build off of FastDVDNet~\cite{tassano2020fastdvdnet}, which is a state-of-the-art video denoiser that implicitly handles motion estimation within the network. We modify this network by replacing the U-Net denoising blocks with an HRNet from~\cite{sun2019deep}, which we find leads to better temporal consistency across our denoised video than the original U-Net architecture. Our denoiser operates on RAW video sequences, Fig.~\ref{fig:denoiser}, and off-the-shelf post-processing is used produce the final output. See Suppl.\ for our full denoiser network architecture and our evaluation against the original FastDVDNet architecture.

\subsection{Training}
We train the denoiser on a combination of synthetic noisy video clips and real still images from camera. First, we pre-train our network for 500 epochs using a combination of real paired stills, synthetic noisy clips from our camera, and synthetic noisy clips from the MOT dataset to help prevent overfitting. After pretraining, we refine the model for 817 epochs on our real still images and synthetic clips from our camera. All images are cropped to 512$\times$512 patches throughout training. We use a combination a perceptual loss (LPIPS)~\cite{zhang2018unreasonable} with an $\lL_1$ loss for our training objective, choosing only the first 3 RAW channels for the LPIPS loss, which requires a 3-channel image. We gamma correct our ground truth images with gamma $=(1/2.2)$, thereby training the denoising network to output a gamma-corrected image. We found that this outperformed applying the gamma correction after denoising. For both pre-training and refinement, we utilize the Adam optimizer~\cite{kingma2014adam} with learning rate 1e-4 and all default parameters. 

\subsection{Post-processing}
Our denoiser is trained on the raw images from the camera. In doing so, this system can work with a number of different post-processing pipelines. To display our final images, we apply the following post-processing steps: demosaicing via bilinear interpolation, white balancing, and histogram equalization. We note that our denoised images are already in a gamma-corrected space. We display the RGB channels of the video clips, omitting the NIR channel in our visualization. We expect that manual post-processing in Adobe Lightroom or a comparable platform could further improve the contrast and perceptual quality of our images.

\section{Evaluation}
\label{sec:results}
First, we evaluate our noise generator performance against several existing noise models for low-light imaging, as well as perform an ablation analysis on the components of our noise model. Next, we compare our noise generator + video denoiser pipeline against several existing denoising schemes. We quantitatively compare on a held-out dataset of still noisy/clean image pairs. Finally, we qualitatively compare on our submillilux dataset of noisy videos, which do not contain ground-truth labels for a quantitative comparison.

\subsection{Noise generator}

After training our noise generator, we assess its performance on a held-out dataset consisting of 832 128$\times$128 video patches. Each patch has 4 color channels and 5 temporal channels. We calculate the KL divergence between our synthetic noise and the real noisy clips after subtracting the clean image. We compare against a non-deep low-light noise model (ELD)~\cite{wei2020physics}, as well as two deep-learning-based noise models, CA-GAN~\cite{chang2020learning} and Noise Flow~\cite{abdelhamed2019noise}. ELD is hand-calibrated using dark frames and grayscale frames to fit to several distributions for different noise sources. Following this calibration scheme, we found that our noise distributions are very different from those described in~\cite{wei2020physics}, likely due to our extremely high gain settings, predominant fixed pattern noise, and periodic components, leading to poor performance of this model. CA-GAN is a camera-aware noise model that takes in a clean image, an estimated shot and read noise image, as well as an example real noisy image from the camera in order to synthesize a signal-dependent noisy image. We use this model off-the-shelf, but find that it generalizes poorly to our camera and noise. Similarly, Noise Flow is designed to work with multiple gain settings and lighting conditions, but does not generalize to our extreme low light, high gain setting. We summarize our findings in Table~\ref{tab:noise_ablation}, and show an example synthetic noisy patch from each method in Fig.~\ref{fig:noise_comparison}. We can clearly see that both Noise Flow and CA-GAN miss the significant banding noise (column offsets) present in our real noisy clips. ELD captures the banding noise pattern well, but misses other components of the noise and does match the real noisy clips visually or in terms of KL divergence. 
 \vspace{-2mm}
\begin{table}[thb]
    \centering
    \small
    \newcolumntype{Z}{S[table-format=1.3,table-auto-round]}
	\begin{tabularx}{\linewidth}{X@{\hspace{5mm}}Z}
    \toprule
    Noise model & {KLD}\\
    \midrule
    
    ELD~\cite{wei2020physics}      & 1.3607 \\
    Noise Flow~\cite{abdelhamed2019noise}       & 0.3855 \\ 
    CA-GAN model~\cite{chang2020learning}       & 0.5130 \\ \midrule
    Ours (ablation)\\\midrule
    $N_{r}$ (Gaussian)                                       & 0.4000 \\
    $N_{s}+N_{r}$  (shot + read noise)                       & 0.4001 \\
    $N_{s}+N_{r}+N_{q}$                                      & 0.1220 \\
    $N_{s}+N_{r}+N_{q} + N_{row}+ N_{row_t}$                 & 0.1184 \\
    $N_{s}+N_{r}+N_{q} + N_{row} + N_{row,t}+N_{p}$          & 0.1133 \\
    $N_{s}+N_{r}+N_{q} + N_{row} + N_{row,t} +N_{p} +N_{f}$  & 0.1375 \\
    $N_{s}+N_{r}+N_{q} + N_{row} + N_{row,t} +N_{p} +N_{f*}$ & 0.0842 \\
    \midrule
    Full model:                                              & \bfseries 0.0691 \\
    \bottomrule
    \end{tabularx}
    \vspace{.5mm}
    \caption{We compare our noise generator to prior work, representative of different approaches to modeling noise distributions. Our method significantly outperforms all baselines. We also present an ablation of components modeled by our noise generators. See Figure~\ref{fig:noise_comparison} for a visual comparison.}
    \label{tab:noise_ablation}
    \vspace{-4mm}
\end{table}

\vspace{1mm}\noindent\textbf{Ablation of noise parameters.}
We ablate different noise components of our generator in Table~\ref{tab:noise_ablation} and show a qualitative comparison in Figure~\ref{fig:noise_comparison}. As before, we calculate the KL divergence between synthetic and real noisy patches, and we find that each component of our noise model improves the KL divergence. Specifically, shot, read, quantization, and row noise, were all used ELD~\cite{wei2020physics}, but were hand-calibrated. Here, we automatically calibrate the different noise components through our GAN training, resulting in better performance than the hand-calibrated model. In addition, our model takes into account the noise behavior over time by including components that are constant over multiple image patches (temporal row noise, fixed pattern noise). As seen in the 5-images averages on the bottom of Figure~\ref{fig:noise_comparison}, our noise model closely matches the average noise, which is important for video denoising. Adding a periodic noise component makes our noise better match the Fourier spectrum of the real noise, which has several prominent peaks in Fourier space (see Suppl.\ for details).  Adding our measured fixed pattern, $N_f$, improves the behavior over time, and learning the fixed pattern, $N_{f*}$ further improves the KL divergence, but at the price of risking overfitting since this is a pixelwise addition of an image to the synthetic noise. In our final model, we use a measured fixed pattern and a learned U-Net which can account for noise that we do not specifically model, such as chromatic effects, or enhance our Gaussian noise approximations to better match the true noise distributions. Our final noise model produces synthetic noise that closely matches the real noise for a single noise instance, over time, and in Fourier space.

\subsection{Full pipeline: video denoising}
Next, we evaluate our video denoiser trained using combination real and synthetic noisy samples against existing denoisers. First, we quantitatively compare against several alternative methods using our dataset of 21 still clean/noisy bursts (since we do not have ground truth for our noisy video clips). We split up our comparison into two categories: single-image denoising methods and video denoising methods, which take in multiple clips at a time.

\begin{table}[htbb]
\centering
\small
\begin{tabular}{|l|l|l|l|}
\hline
Method      & PSNR & SSIM & LPIPS\\ \hline
\textbf{Single Image Methods:} & & & \\ \hline
Noise2Self~\cite{batson2019noise2self} & 20.11 & 0.210 &  0.545 \\ \hline 
Unprocesing \cite{brooks2019unprocessing} & 12.86 & 0.249 & 0.355 \\ \hline 
L2SID (pretrained)~\cite{chen2018learning}      & 13.6 & 0.512& 0.338\\ \hline
L2SID (retrained) ~\cite{chen2018learning}    & 26.9 & 0.892 & 0.198 \\\hline
\textbf{Video Methods:} & & & \\ \hline
V-BM4D ~\cite{maggioni2012video}     & 16.2  & 0.322  & 0.419  \\\hline
pretrained PaCNet~\cite{vaksman2021patch}& 13.65 & 0.512  & 0.338 \\\hline
pretrained FastDVDnet~\cite{tassano2020fastdvdnet}& 23.8 &  0.618  & 0.282 \\\hline
\textbf{ours}        & \textbf{27.7}  & \textbf{0.931}   & \textbf{0.078}  \\\hline
\end{tabular}
\vspace{.5mm}
\caption{Performance on still images from test set.}
\label{tab:results_stills}
\vspace{-4mm}
\end{table}

For single-image denoising, we compare against two pretrained deep denoising methods: Unprocessing~\cite{brooks2019unprocessing}, which operates on RAW images and is trained using different read and shot noise levels, as well as L2SID~\cite{chen2018learning} which takes in a raw noisy image and jointly denoises and processes the image. Both of these methods do not perform well on our dataset, due to the extreme noise in our raw measurements. We also retrained L2SID~\cite{chen2018learning} using our still image pairs, resulting in better performance single-images, but significant flickering over time for video (see Suppl.\ for example). Noise2Self, a self-supervised approach, does poorly with our noise due to its highly structured content (e.g.\ correlated lines for the row offsets), and results in denoised images with prominent line artifacts. These results are summarized in Table~\ref{tab:results_stills}, with full images shown in the Suppl. 

\begin{figure*}[thb]
\centering
\includegraphics[width=\textwidth]{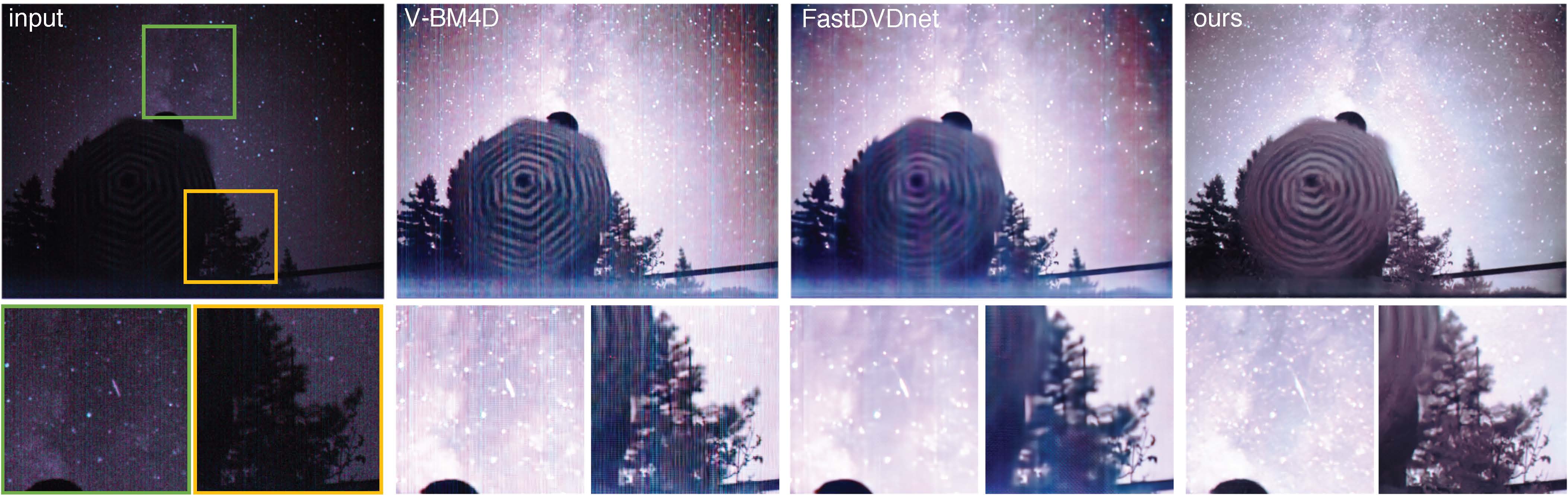}
   \caption{Results on noisy video clips taken at 10 fps in 0.0006~\si{\lux}. The input sequence (left), V-BM4D, pretrained FastDVDnet, and our results are shown. Our method maintains more details throughout the clip and does not contain the prevalent streaking artifacts that are present in V-BM4D and the pretrained FastDVDnet. See supplement for full video clips. (Digital zoom recommended.)}
\label{fig:results}
\vspace{-4mm}
\end{figure*}

\begin{figure}[thb]
\centering
\includegraphics[width=\linewidth]{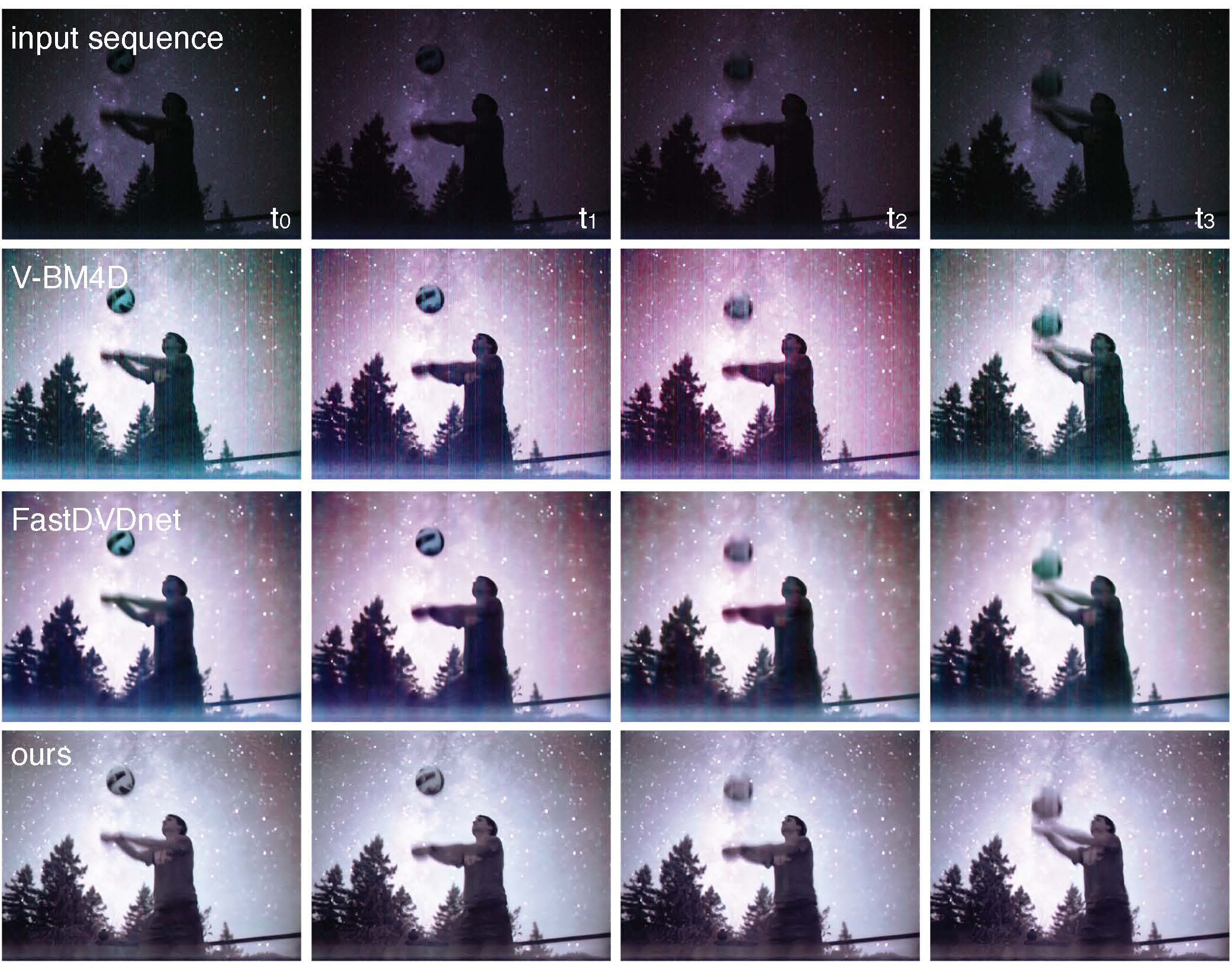}
   \caption{Results on noisy video clip, showing performance over time for a video taken at 10 fps in 0.6~\si{\milli\lux}. Our method achieves better temporal consistency than V-BM4D or the pretrained FastDVDnet, and also has fewer artifacts within each frame. See supplement for full video clips. (Digital zoom recommended.)}
\label{fig:results_motion}
\vspace{-8mm}
\end{figure}

For video denoising, we feed in 5 noisy clips to each denoiser, then compare against a single still ground-truth image. We compare our method against V-BM4D (a classic video denoising method), as well as two pre-trained state of the art deep video denoisers, FastDVDnet~\cite{tassano2020fastdvdnet} and PaCNet~\cite{vaksman2021patch}. Both of these models use additive Gaussian noise, so as expected, they do not perform well for our real noisy clips. FastDVDnet, which is designed to operate at multiple noise levels, outperforms PaCNet~\cite{vaksman2021patch}, which is designed for a specific Gaussian noise level. Our denoising method, which is based on a modified FastDVDnet and trained using our noise generator, achieves the best performance. This demonstrates the importance of having a realistic noise model during denoiser training.


Next, we qualitatively compare our performance on our unlabeled dataset of submillilux video clips. We show the performance of our method as compared to V-BM4D and the pretrained FastDVDnet in Figure~\ref{fig:results}, with additional video comparisons available in the Suppl. Our method had fewer horizontal streak artifacts than the other methods, maintains more details such as stars, and has better overall image quality. We can clearly see the Milky Way in the denoised videos, and our method is robust to fast-moving objects in the background (e.g.\ we capture a shooting star in Fig.~\ref{fig:results}). When viewing the adjacent frames in the video clip, Fig.~\ref{fig:results_motion}, our method has less flickering than V-BM4D and the pretrained FastDVDnet, which both have significant flickering between frames, likely due to the significant noise present in the raw data. 

Finally, we perform a perceptual experiment with blind randomized A/B tests between our method, V-BM4D, FastDVDNet, and L2SID using 10 clips from our video dataset. Throughout 300 comparisons with 10 workers, our method is rated as having superior image quality than the alternative methods over 95\% of the time (details in Suppl.).

\section{Conclusion and Discussion}
We have demonstrated photorealistic video denoising at submillilux levels of illumination for the first time. We achieved this through a combination of excellent camera hardware (a low-light optimized RGB+NIR camera), a physics-inspired noise generator used to generate realistic noisy video clips, and a video denoiser trained using a combination of real still images and synthetic noisy video clips. Our work showcases the power of deep-learning-based denoising for extremely low-light settings. We hope that this work leads to future scientific discoveries in extremely low light levels (e.g. studying nocturnal animal behavior in moonless conditions or under a forest canopy), and will help push the limits of robot vision in extremely dark settings. Potential misuse of this work includes night-time surveillance or use in conjunction with weapons systems. 

Our approach has limitations. First, our noise generator is limited to producing noise that mimics a single gain setting (in this case, the highest gain). Future work could expand the noise generator model to work with multiple camera gains/ISOs. Next, our denoised night videos have muffled colors due to the dominance of NIR over RGB at night. Work in style transfer and recoloring could further improve the visual appearance of the denoised video clips by enhancing embedded color cues or synthesizing realistic-looking colors. Finally, the performance of the denoiser may be improved in the future through class-aware denoising~\cite{remez2018class} and joint denoising/segmentation.


{\small
\bibliographystyle{ieee_fullname}
\bibliography{egbib}
}

\end{document}